# Two-Stream Binocular Network: Accurate Near Field Finger Detection Based On Binocular Images

Yi Wei[1], Guijin Wang[2], Cairong Zhang[3], Hengkai Guo[4], Xinghao Chen[5], Huazhong Yang[6]

*Department of Electronic Engineering, Tsinghua University, Beijing 100084, China*

{[1]wei-y15, [3]zcr17, [4]guohk14, [5]chen-xh13}@mails.tsinghua.edu.cn

{[2]wangguijin, [6]yanghz}@tsinghua.edu.cn

*Abstract*— Fingertip detection plays an important role in human computer interaction. Previous works transform binocular images into depth images. Then depth-based hand pose estimation methods are used to predict 3D positions of fingertips. Different from previous works, we propose a new framework, named Two-Stream Binocular Network (TSBnet) to detect fingertips from binocular images directly. TSBnet first shares convolutional layers for low level features of right and left images. Then it extracts high level features in two-stream convolutional networks separately. Further, we add a new layer: binocular distance measurement layer to improve performance of our model. To verify our scheme, we build a binocular hand image dataset, containing about 117k pairs of images in training set and 10k pairs of images in test set. Our methods achieve an average error of 10.9mm on our test set, outperforming previous work by 5.9mm (relatively 35.1%).

## I. INTRODUCTION

In recent years, hand pose estimation, especially fingertip detection, has drawn lots of attention from researchers because of its application on human computer interaction and augmented reality. Traditional hand pose estimation based on binocular vision first converts binocular images into depth images [13]. Then some popular depth-based hand pose estimation methods [1][5][6] can be applied. However, these methods depend on high precision of depth data. Chen proposes a method to predict 3D positions of fingertips and palm root from binocular images directly [12]. However, their method has poor performance on complex datasets.

Previous hand pose estimation methods can't work well on fingertip detection. As proposed in [5], fingertip estimation is harder than other hand joints due to large range of motion and imprecise depth data. It's meaningful to research into fingertip detection because fingertips play a more important role in human computer interaction and augmented reality than other joints.

In this paper, we propose a new method based on deep convolutional networks detecting fingertips directly from binocular images. Original images and mask images are used as two channels of inputs. We argue that right and left images have similar low level features. Thus TSBnet first shares convolutional layers for right and left images. Then high level features of left and right images in two-stream convolutional networks are extracted separately. Further, we add a new layer called binocular distance measurement layer, which can transform pixel coordinates of fingertips to their 3D positions. Compared with Chen's method [12], our approach can achieve higher precision. To verify our method, we build a new binocular hand image dataset. Our dataset is more complex, large and diverse. It contains 8 people's hand images, 16 basic hand poses and some extra ones. Our training set has 117k pairs of binocular images while test set has 10k pairs of binocular images.

## II. DATASET

In this section we will describe our dataset of binocular hand images for fingertip detection. There are mostly 4 publicly datasets for hand pose estimation: NYU [16], ICVL [14], MSRA [1] and HandNet [17]. However, all these datasets are based on depth images. [12] proposed a dataset of binocular hand images, but their dataset is small and has only one person's hand images. In the meanwhile, their test set are selected from all images randomly, which has high similarity with training set so cannot test the generalization of the model.

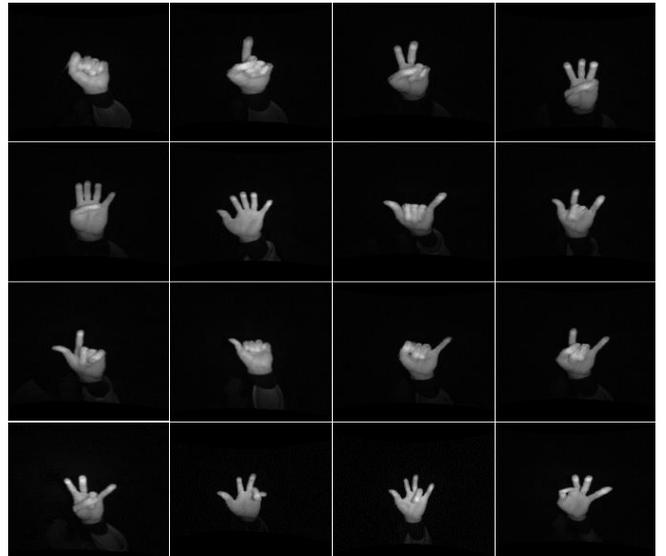

Fig. 1 Basic poses in our dataset

For these reasons, we build a new dataset. We use leap motion to capture binocular hand images with the resolution of 640x480, and use trakstar[10] to get 3D positions of five fingertips and palm root. Trakstar [10] is a kind of DC magnetic tracker which can provide 1.4mm accuracy for 3D position of objects.

Totally, our training set has about 117k pairs of binocular hand images from 8 people's hand with 16 kinds of basic hand poses as shown in Fig.1, and some extra ones. The test set has 10k pairs of binocular hand images, collected from 2 people. Some hand poses in test set have a large range of rotation and motion, which can test our model's generalization ability.

## III. TWO-STREAM BINOCULAR NETWORK

In this section we will describe our algorithm based on binocular images for fingertip detection in details.

### A. Baseline Network Architecture

This part introduces our baseline network architecture (See Fig.2), which uses a pair of binocular images as input. We propose that a pair of binocular images have similar low level features. The shared convolutional layers extract low level features of right and left images together. Then the low level feature maps of left image are fed into one stream ConvNet while the ones of right image are fed into another. Fully connected layers fuse information from two streams. The output of fully connected layers is an 18-element vector with the form $((u_l + u_r)/2, u_l - u_r, (v_l + v_r)/2)$ of five fingertips and palm root. $(u,v)$ means pixel position of tips. Index l, r mean left and right images. In our dataset, $v_l$ equals to $v_r$.

Finally we use binocular distance measurement principle to transform pixel position $((u_l + u_r)/2, u_l - u_r, (v_l + v_r)/2)$ to 3D position $(x, y, z)$.

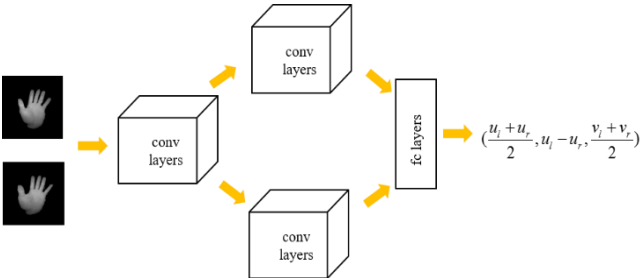

Fig. 2 Our baseline network architecture

### B. The Whole Architecture of TSBnet

Fig.3 shows our novel architecture to improve baseline. TSBnet uses original and mask images as input to enhance the model robustness. Further, we embed binocular measurement layer into network architecture so that we can achieve end-to-end learning.

*1) Use Original Images and Mask Images as Input:* Mask images are robust to variation of illumination and skin colors, so they can enhance the robustness of our model. Original images contain more information of the hand so the model can learn more from them. We use both original images and mask images as input, so the input of TSBnet are four images: mask and original image of left and right. We use the method proposed in [12] to extract mask images.

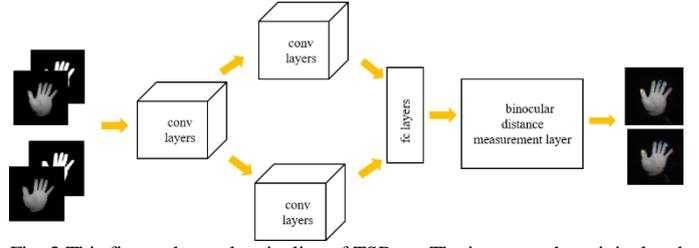

Fig. 3 This figure shows the pipeline of TSBnet. The input are the original and mask images of right and left. Fully connected layers fuse left and right information. Finally binocular distance measurement layer's outputs are 3D positions of five fingertips and palm root.

*2) Binocular Measurement Layer:* The baseline mentioned in A requires a post processing stage. We create a new layer following binocular distance measurement principle to avoid post processing. In this way, we can train our model end-to-end. What we finally predict is the 3D position $(x, y, z)$ of fingertips in a coordinate whose origin is the centre of left and right cameras. $(u_l, v_l)$ and $(u_r, v_r)$ are positions (in pixels) of a fingertip in left and right images. By multiplying a factor $\lambda$, we can get fingertip positions in millimetres $(x_l, y_l)$ and $(x_r, y_r)$. The disparity can be calculated as $d = x_l - x_r$. Then we can get $(x, y, z)$ by the following equations.

$$x = (\frac{u_l + u_r - w}{2})\frac{\lambda z}{f}$$
$$y = (\frac{v_l + v_r - h}{2})\frac{\lambda z}{f} \qquad (1)$$
$$z = \frac{fb}{d} = \frac{fb}{\lambda(u_l - u_r)}$$

Where $f$ is the camera intrinsic, $b$ is the displacement of the binocular cameras, $w$ is the width of images and $h$ is the height of images.

When using back propagation algorithm for training, we must calculate gradient. See the following equations.

$$\frac{\partial L}{\partial(\frac{u_l + u_r}{2})} = \frac{\partial L}{\partial x} \cdot \frac{\lambda z}{f}$$
$$\frac{\partial L}{\partial(\frac{v_l + v_r}{2})} = \frac{\partial L}{\partial y} \cdot \frac{\lambda z}{f} \qquad (2)$$
$$\frac{\partial L}{\partial(u_l - u_r)} = \frac{\partial L}{\partial x} \cdot (-\frac{xz\lambda}{fb}) + \frac{\partial L}{\partial y} \cdot (-\frac{yz\lambda}{fb}) + \frac{\partial L}{\partial z} \cdot (-\frac{z^2 \lambda}{fb})$$

Where L is the loss function.

### C. Implementation Details

In this section, we explore some techniques used in TSBnet. We also introduce a new training method to train TSBnet.

*1) Multi-scale Training for TSBnet*: To increase robustness of our model, we use multi-scale training. The original images of input are cropped from a 640x480 image centred on the centroid of hand with different sizes: 240x240, 200x200 and

160x160. Mask images also have 3 different sizes. Then all original and mask images will be resized into 96x96 for input.

*2) More channels:* We double channel numbers of every convolutional layer. Different channels can extract different kinds of features from images so we can enrich features.

*3) Residual Connection:* To improve the learning ability of TSBnet, we add two residual connections [7] in our network architecture (See Fig.4). Residual connection can be explained as the ensembles of shallow networks [3] and it can improve model's performance.

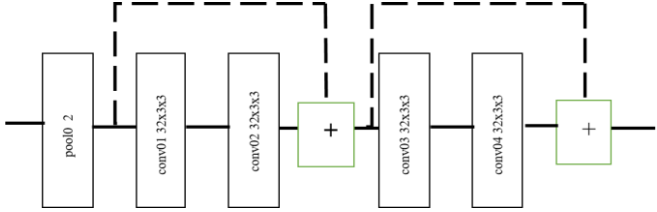

Fig. 4 Residual connection in TSBnet

*4) New Training Method:* After adding binocular distance measurement layer, we can train the model end-to-end. However, if we train the model without a good initialization, exploding gradient problem is easy to take place. We argue the reason for exploding gradient problem is that in Eq(2), the factor $\lambda z/f$ will enlarge the gradient when using back-propagation algorithm. So first we pre-train a model whose label is $((u_l+u_r)/2, u_l-u_r, (v_l+v_r)/2)$ (See Fig.5). This pre-trained model can get a precise prediction of $((u_l+u_r)/2, u_l-u_r, (v_l+v_r)/2)$. And through binocular distance measurement layer, it will make a precise prediction of $(x,y,z)$. When fine-tuning on this pre-trained model end-to-end with binocular distance measurement layer, exploding gradient problem will not occur and we will get a preciser model (See Fig.5).

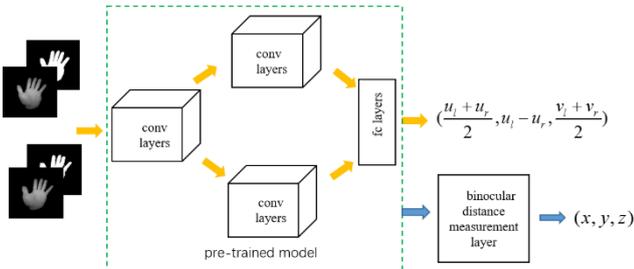

Fig.5 New training method: First we need to pre-train a model. The model's outputs are $((u_l+u_r)/2, u_l-u_r, (v_l+v_r)/2)$ of five fingertips and palm root. Then we finetune the pre-trained model end-to-end.

Here comes out the whole architecture of the TSBnet (See Fig.3). Compared with the baseline mentioned in A, TSBnet has residual connection, more channels and binocular distance measurement layer and use multi-scale training. The input of TSBnet are four images: original and mask images both of left and right, while the output are 3D positions of five fingertips and palm root. Table 1 shows the whole architecture of TSBnet, where $conv, d\times(s\times s)\times k$ means k convolutional layers with the filter size of s and d feature channels, $pool, s\times s$ means a max-pooling layer with the kernel size of s, and $fc, n$ means a fully connected layer with n neurons. There is a PReLU (parametric Rectified Linear Unit) layer after every convolutional layer and fully connected layer.

TABLE I
THE WHOLE ARCHITECTURE OF TSBNET

| Left | Right | |
|---|---|---|
| conv0, 32x(5x5) | conv0, 32x(5x5) | shared |
| pool0, 2x2 | pool0, 2x2 | |
| conv0x, 32x(3x3)x4 | conv0x, 32x(3x3)x4 | shared |
| pool1, 2x2 | pool1, 2x2 | |
| conv1x, 48x(3x3)x2 | conv1x, 48x(3x3)x2 | shared |
| conv2x, 64x(3x3)x3 | conv2x, 64x(3x3)x3 | separate |
| pool2, 2x2 | pool2, 2x2 | |
| conv31, 128x(3x3) | conv31, 128x(3x3) | separate |
| conv32, 192x(3x3) | conv32, 192x(3x3) | separate |
| fc1, 512 | | |
| fc2, 256 | | |
| fc3, 18 | | |
| binocular distance measurement layer | | |

## IV. EXPERIMENTS

In this section we will show the performance of TSBnet, including comparison with previous work and exploration study of refinement in every modification on network architecture mentioned in section Ⅲ.

### A. Experiment Setup

We train TSBnet with Caffe [4] using C++ on Nvidia GPU Titan X Pascal. We use stochastic gradient descent (SGD) with a mini-batch size of 128. The learning rate starts from 0.005 and decays by 10 times every 200000 iterations. The weight decay is 0.0005 and momentum is 0.9.

We use two metrics to evaluate the performance: 1) Average 3D distance error of each joint, defined as Euclidean distance of truths and our predictions. 2) Percentage of success frames, defined as the percentage of frames where 3D distance errors of all joints are below a threshold.

### B. Comparison with Previous work

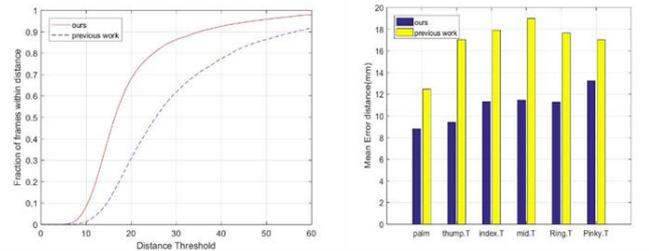

Fig.6 Comparison with method proposed in [12] on our dataset: distance error (left) and percentage of success frames (right).

Task in [12] is similar to ours and we can compare our method with theirs. Our method can achieve a mean error for all joints of 10.9mm on our dataset while theirs is 16.8mm. Fig.6 shows that our method performs much better than theirs.

Leap motion has its own algorithm to predict 3D position of fingertips but it is not published. However, qualitatively, we can show comparison between its algorithm and ours. Fig.7 shows some images in our test set predicted by our method and theirs. Their method performs worse when fingers bend. In human-computer interaction, we usually bend fingers to imitate click. So this kind of hand pose often appears and we need a precise prediction of it.

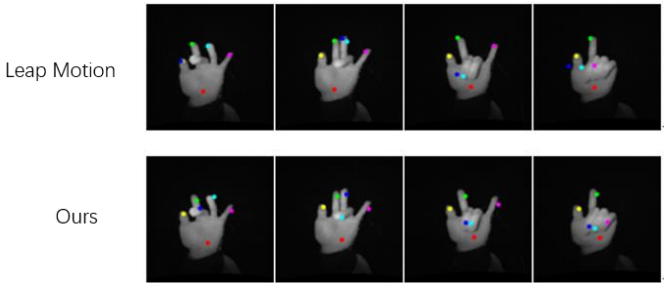

Fig.7 The first line is the prediction of leap motion and the second line is ours.

## C. Module Analysis

In this section, we focus on the investigation of good practices. We introduce five modifications on our baseline network. The baseline is a two-stream network without residual connection, binocular distance measurement layer and multi-scale training, whose inputs are just original images. We introduce our modifications gradually: 1) multi-scale training to improve the robustness of our network 2) increase channels of convolutional layers to enrich features 3) use both original and mask images as input to balance information details and model robustness 4) residual connections 5) binocular distance measurement layer, so our model is end-to-end. Fig.8 and Table 2 are the experiment results. All strategies reduce mean error by 2.7mm (relatively 19.9%) totally.

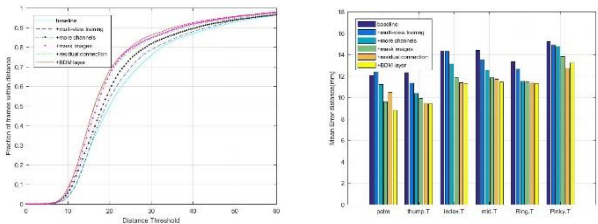

Fig.8 Module analysis on our dataset: distance error (left) and percentage of success frames (right). "+BDM layer" means "+binocular distance measurement layer".

TABLE II
MEAN ERRORS OF INCREMENTAL STRATEGIES ON OUR DATASET. LOWER IS BETTER.

| Strategy | Mean error(mm) |
|---|---|
| baseline | 13.6 |
| +multi-scale training | 13.2 |
| +more channels | 12.3 |
| +mask images | 11.4 |
| +residual | 11.2 |
| +binocular distance measurement layer | 10.9 |

## V. CONCLUSIONS

We propose a new network named TSBnet to predict 3D positions of five fingertips and palm root from binocular images. It is a Two-Stream CNN connected with binocular distance measurement layer. We build a new binocular hand images dataset to demonstrate our method, whose training set has 117k pairs of left and right images and test set has 10k pairs of images. Compared with previous work, our TSBnet has achieved higher accuracy on our dataset.

ACKNOWLEDGMENT

This work is supported by State High-Tech Development Plan of China (863 Program, No.2015AA016304).